\documentclass{article}
\usepackage{ijcai23}

\usepackage{times}
\usepackage{soul}
\usepackage{url}
\usepackage[hidelinks]{hyperref}
\usepackage[utf8]{inputenc}
\usepackage[small]{caption}
\usepackage{graphicx}
\usepackage{amsmath}
\usepackage{amsthm}
\usepackage{booktabs}
\usepackage{algorithm}
\usepackage{float}
\usepackage[switch]{lineno}
\usepackage{amsmath,latexsym,amssymb,amsthm,array,amsfonts,algorithm,booktabs,graphicx,subfigure,multirow,cuted,stfloats}
\usepackage{mathrsfs}
\usepackage{amsfonts,amssymb}
\usepackage{threeparttable}
\usepackage{fancyvrb}
\usepackage{tcolorbox}
\usepackage{listings}
\usepackage{amsmath}
\usepackage{algpseudocode}

\title{Federated Incomplete Multi-View Clustering with \\ Heterogeneous Graph Neural Networks}

\author{
Xueming Yan$^{1,3}$
\and
Ziqi Wang$^{2}$
\and
Yaochu Jin$^{3*}$\and
\affiliations
$^1$Guangdong University of Foreign Studies\\
$^2$ East China University of Science and Technology\\
$^3$School of Engineering, Westlake University\\
\emails
yanxm@gdufs.edu.cn,
wzq\_7379@163.com,
jinyaochu@westlake.edu.cn
}
\begin{document}
\maketitle
\begin{abstract}
Federated multi-view clustering offers the potential to develop a global clustering model using data distributed across multiple devices. However, current methods face challenges due to the absence of label information and the paramount importance of data privacy. A significant issue is the feature heterogeneity across multi-view data, which complicates the effective mining of complementary clustering information. Additionally, the inherent incompleteness of multi-view data in a distributed setting can further complicate the clustering process.
To address these challenges, we introduce a federated incomplete multi-view clustering framework with heterogeneous graph neural networks (FIM-GNNs). 
In the proposed FIM-GNNs, autoencoders built on heterogeneous graph neural network models are employed for feature extraction of multi-view data at each client site. At the server level, heterogeneous features from overlapping samples of each client are aggregated into a global feature representation. Global pseudo-labels are generated at the server to enhance the handling of incomplete view data, where these labels serve as a guide for integrating and refining the clustering process across different data views.
Comprehensive experiments have been conducted on public benchmark datasets to verify the performance of the proposed FIM-GNNs in comparison with state-of-the-art algorithms.
\end{abstract}

\section{Introduction}
Multi-view clustering is a fundamental machine learning task aiming to improve clustering performance by leveraging the consistency and complementary information from different views \cite{wen2020generalized,cai2024multiview,yan2022multimodal}. Traditional multi-view clustering methods typically process raw data directly or through basic feature transformations, but the performance of these approaches tend to deteriorate on high-dimensional data \cite{zhong2021improved,zhong2022multi}. To overcome these limitations, deep learning technologies have been introduced into multi-view clustering. By utilizing neural networks to extract complex and abstract features from views, deep multi-view clustering methods are capable of handling more complex data structures and relationships \cite{li2023multi,yan2024binary}.

Despite significant achievements of deep multi-view clustering methods in dealing with traditional Euclidean space data \cite{ren2022deep,xiao2023dual}, real-world data often naturally exists in graph forms, such as social networks and knowledge graphs, presenting new challenges for clustering tasks. In particular, the complexity of graph structures makes it difficult for traditional clustering methods to be effective \cite{yan2019graph}. Graph neural networks (GNNs) have gained widespread attention for their ability to capture structural information effectively \cite{wang2019gmc}. In multi-view clustering applications \cite{cheng2021multi,wang2019attributed}, GNNs can utilize the relationships between views as well as structural and feature information to obtain node features suited for clustering, effectively integrating multi-view data and enhancing clustering performance. However, existing methods based on GNNs often overlook the heterogeneity of data across different views, and the node features obtained may be unsuited for clustering outcomes \cite{fan2020one2multi}. Additionally, these methods typically employ centralized training, which can raise privacy concerns of multi-view data \cite{guo2022graph}.

\begin{figure}[htp]
    \centering
    \includegraphics[width=8cm]{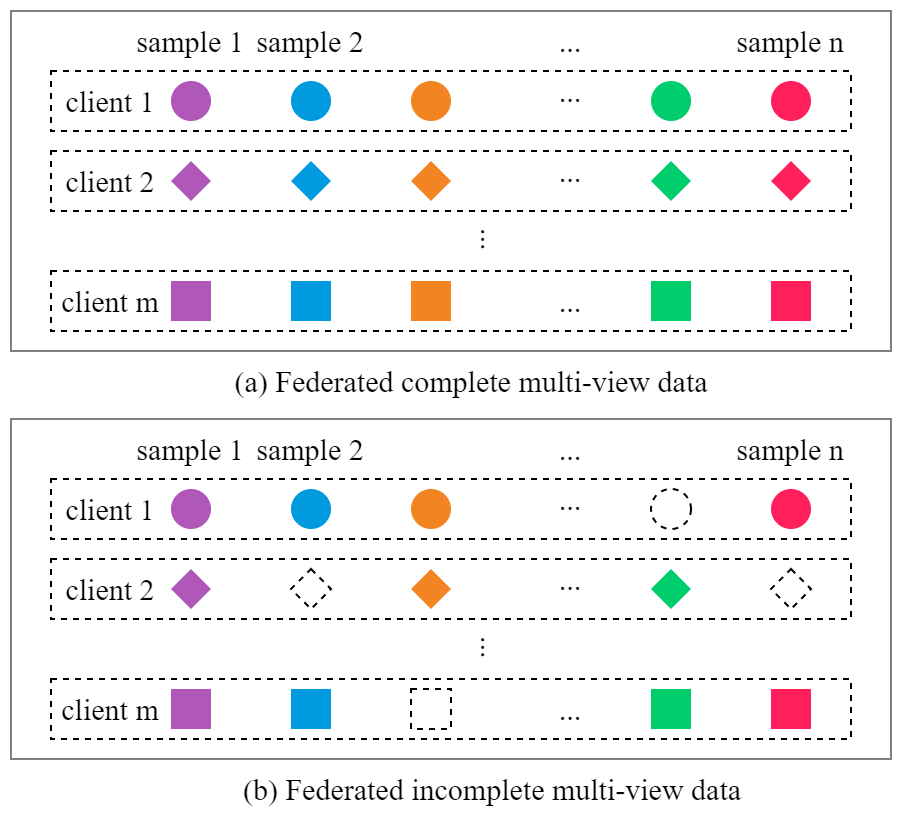}
    \caption{The disparity between complete and incomplete multi-view federated data. (a) Federated complete multi-view clustering, where different clients possess complete, distinct sets of sample features. (b) Federated incomplete multi-view clustering, where different clients possess different sets of sample features. Each client may have missing samples, but each sample exists in at least one client.}
    \label{arch11}
\end{figure}

Federated learning, as a distributed learning model, offers a solution to address data heterogeneity issues while protecting data privacy \cite{liu2023secure}. By combining federated learning with graph neural networks, it is possible to capture graph structures and node features while ensuring data privacy\cite{zheng2023federated}. However, federated GNN-based methods face dual challenges in solving clustering problems: inconsistency and heterogeneity of data across different views \cite{huang2022efficient,huang2020federated}. Data from different views are not always perfectly overlapping and vary in data features and sizes. Moreover, due to various uncontrollable factors in practical applications, as shown in Fig. \ref{arch11}, multi-view data may be incomplete, further complicating clustering tasks \cite{ren2024novel,chen2023federated}.

To address these challenges, this paper proposes a federated incomplete multi-view clustering framework based on heterogeneous graph neural networks (FIM-GNNs). This framework aims to capture node features with heterogeneous GNNs from different views, and the global aggregation is designed to merge the complementary information from various views. Additionally, with the help of global pseudo labels, we merge feature extraction and clustering into a unified process, and use these labels to assist client training to achieve consistency in incomplete multi-view data. The main contributions of this paper can be summarized as follows:

\begin{itemize} 
\item  We propose a federated incomplete multi-view clustering framework utilizing heterogeneous GNNs. Local training with heterogeneous GNNs and a global aggregation are introduced to effectively harness the complementary information in multi-view data, significantly enhancing the clustering performance.

\item  A global pseudo-label mechanism is designed with heterogeneous aggregation in a federated environment, enhancing the ability of the FIM-GNNs to handle incomplete view data and improving the consistency of features as well as the performance of the clustering results.

\item Comprehensive experiments demonstrate the competitive performance of the FIM-GNNs in handling data heterogeneity and incompleteness compared to four state-of-the-art incomplete multi-view clustering methods across three public datasets.
\end{itemize}

\section{Related work}
This section provides the problem formulation of federated incomplete multi-view clustering and a short overview of existing multi-view clustering techniques. The notations in this study are listed in Table \ref{tab_lab}.

\subsection{Problem formulation} 
For a multi-view dataset with \(N\) samples, where \(X_c = \{X_c^1, \ldots, X_c^i, \ldots, X_c^m\}\), \(X_c^i \in \mathbb{R}^{N \times D_m}\) represents the feature set of the nodes, and the sample features \(D_i\) of different views exhibit differences.
In practice, some views may have missing sample data. Therefore, the description of an incomplete multi-view clustering dataset is as follows:
\begin{equation}
X^i = (M^i \otimes \mathbf{1}) \odot X_c^{i}
\end{equation}
which is subject to the following constraint:
\begin{equation}
\sum_{i=1}^N M_j^i \geq 1
\end{equation}
Here, \(M = \{M^1, \ldots,M^i, \ldots,M^m\}\) represents a set of binary selection matrices, with \(M^i \in \{0,1\}^{N \times 1}\). The entries of \(M^i = 1\) indicate that the corresponding features are retained, whereas \(M^i = 0\) means that the features are discarded. 
 \(\mathbf{1} \in \mathbb{R}^{1 \times N}\), and \(M^i \otimes \mathbf{1}\) indicates that either all sample features are missing or all are present, meaning that partial missing features within the samples are not considered. \(\odot\) represents the Hadamard product. 
\(\sum_{j=1}^N M_j^i \geq 1\) ensures that each sample retains at least one feature, and no sample is completely devoid of features.

In this study, we consider the problem of incomplete multi-view clustering under a federated framework, as shown in Fig. \ref{arch11}. Assume there is an incomplete multi-view dataset $X = \{X_1, \ldots, X_m\}$ and $A = \{A_1, \ldots, A_m\}$ distributed across $m$ clients. For client $i$, $X_i \in \mathbb{R}^{N_i \times D_i}$ represents the node features, and $A_i \in \mathbb{R}^{N_i \times N_i}$ represents the graph structure. Due to the data heterogeneity in a federated setting, both the sample dimensions $D_i$ and the number of samples $N_i$ vary across clients. Additionally, due to missing data, $N_i < N$. The incomplete multi-view clustering problem involves extracting features from the incomplete node features and graph structures of each view, followed by an aggregation to produce the final clustering results.

\begin{table}[h]
\centering
\scalebox{0.8}{
\begin{tabular}{c|l} 
\toprule
\textbf{Symbols} & \textbf{Descriptions}                 \\ \hline
m                 & Number of clients                       \\ \hline
N                 & Number of samples                     \\ \hline
$D_i$               & Feature dimension of the $i$-th client           \\ \hline
$N_i$               & Number of samples in the $i$-th client             \\ \hline
$X^i$               & Node features of the $i$-th client    \\ \hline
$A^i$               & The graph structure matrix of the $i$-th client          \\ \hline
$Z^i$               & Features extracted from the $i$-th client           \\ \hline
$Z_j^i$             & Features of sample $j$ extracted by the $i$-th client \\ \hline
$Z_C^i$             & Features extracted based on overlapping data from the $i$-th client \\ \hline
$U^m$               & Cluster centers of the m-th view        \\ \hline
Z                 & Global features of overlapping samples                                   \\ \hline
C                 & Global cluster center                                   \\ \hline
P                 & Global pseudo-labels       \\ \hline
$R_{i}$                 & The missing rate of the $i$-th view      \\ \bottomrule
\end{tabular}}
\caption{Summary of the main notations}
\label{tab_lab}
\end{table}

\subsection{Multi-view clustering}
Recently, GNNs have emerged as effective tools for graph clustering tasks \cite{tsitsulin2023graph} to deal with multi-view data. GNN-based graph clustering techniques combine GNNs with autoencoders to perform clustering tasks in an unsupervised manner. For example, O2MAC \cite{fan2020one2multi} captures features using a shared multilayer graph convolutional encoder across different views and performs graph reconstruction with a multilayer graph convolutional decoder in each view. DAEGC \cite{wang2019attributed} utilizes a graph attention-based autoencoder that considers both node attributes and the structural information of the graph. MAGCN \cite{cheng2021multi} extends this approach to multi-view graph clustering, taking into account the geometric relationships and probabilistic distribution consistency between multi-view data to further enhance clustering tasks. GC-VAE\cite{guo2022graph} combines variational graph autoencoders with graph convolutional networks, clustering nodes based on graph topology and node features. However, these methods utilize centralized training and overlook the importance of data privacy protection.

GNNs-based multi-view clustering aims to enhance clustering performance by leveraging consistent and complementary information across multiple views \cite{hu2023efficient}. Meanwhile, federated multi-view clustering provides a promising solution for protecting data privacy  in various distributed devices/silos, as discussed in \cite{qiao2024federated}.
For example, a distributed multi-view spectral clustering method called FMSC \cite{wang2020federated} employs homomorphic encryption and differential privacy techniques to protect data privacy. However, the time-consuming encryption and decryption processes reduce the efficiency of the model. By considering the communication issues, FedMVL \cite{huang2022efficient} proposes a vertically federated learning framework based on non-negative orthogonal decomposition, effectively reducing the cost of the model. However, these shallow methods struggle to effectively extract node information, limiting model performance. FL-MV-DSSM \cite{huang2020federated} utilizes a deep structured semantic model to construct a multi-perspective recommendation framework. However, these federated learning multi-view clustering methods overlook the incompleteness of data across different views and are limited by their clustering approaches, which constrains model performance. 

Recently, some researchers \cite{xu2022deep,wen2021structural,lin2021completer} have started paying attention to incomplete clustering algorithms, which employ the complete views to predict the missing data. For example, EE-IMVC \cite{liu2020efficient} imputed each incomplete base matrix generated by incomplete views with a learned consensus clustering matrix. Moreover, IMVC-CBG \cite{wang2022highly} reconstructed missing views with a consensus bipartite graph by minimizing the conditional entropy of multiple views using dual prediction. After that, IMVC \cite{xu2023adaptive} learns the features for each view by autoencoders and utilizes an adaptive feature projection to avoid the imputation for missing data. Fed-DMVC \cite{chen2023federated} addresses the incompleteness issue in federated multi-view clustering with an autoencoder model. However, these methods do not simultaneously consider both data heterogeneity and data incompleteness issues in the process of multi-view clustering.

\section{Proposed Method}
\begin{figure*}[htbp]
    \centering
    \includegraphics[width=18cm]{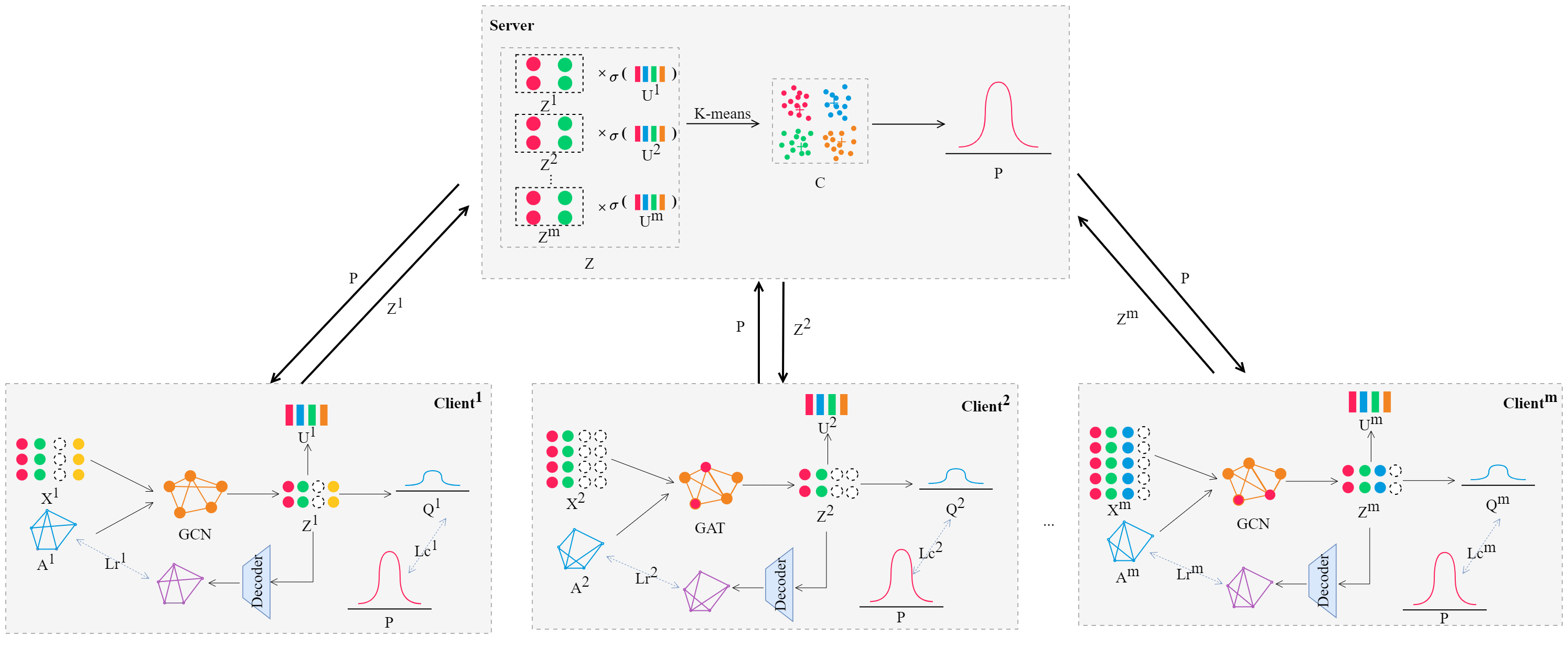}
    \caption{Overview of FIM-GNNs. There are $m$ clients and one server. Initially, the clients perform feature extraction using GAT or GCN based on local features. We employ a decoder for graph reconstruction and the global pseudo-label $P$ as auxiliary training information.}
    \label{arch12}
\end{figure*}

In this section, we present the federated incomplete multi-view clustering with heterogeneous graph neural networks (FIM-GNNs), illustrated in Figure \ref{arch12}. 
Heterogeneous graph neural networks, such as GCN \cite{zhang2019graph} and GAT \cite{velivckovic2017graph}, are utilized to extract features and tailored for local client-side training. These incomplete multi-view features are then combined at the server to generate global pseudo-labels. Following this, clustering results are derived through the algorithm optimization. Lastly, we provide an analysis of the time complexity for the proposed FIM-GNNs.

\subsection{Local training with heterogeneous GNNs}
Based on the heterogeneity of the multi-view data at each client side, we utilize the different GNNs as graph autoencoders for feature extraction in different view data. In addition, we introduce global pseudo-labels \( P \) acquired from the server as labels to assist client-side training.
In the graph autoencoder process, we extract low-dimensional features from the client side, aiming to capture the latent features of the multi-view data while preserving data privacy.
For the \( i \)th client 's data \( X^i \) and graph structure \( A^i \), we use the GNN-based encoder method to project them into low-dimensional features \( Z^i \), and utilize a decoder to reconstruct the graph structure \( A^i\). Considering the differences in sample quantity and dimension across different views (clients), we adopt two types of GNN models, GCN and GAT, to construct graph autoencoders for feature extraction.
When the missing rate of the $i$-th view data is not too low, we construct a two-layer GCN as the encoder for feature extraction:
\begin{equation}
 Z = f(X, A) = \text{softmax}\left(\hat{A}\text{ReLU}\left(\hat{A} X W^{(0)}\right) W^{(1)}\right) 
\end{equation}
where \(\hat{A} = \tilde{D}^{-1/2} \tilde{A} \tilde{D}^{-1/2}\), and \(W^{(0)}\), \(W^{(1)}\) are the parameters for the first and second layers of the GCN, respectively.
When the missing rate of the $i$-th view data is low, we use a two-layer GAT as the encoder for feature extraction:
\begin{equation}
z_j^{(1)} = \sigma \left( \sum_{k \in N_j} \alpha_{jk} W^{(0)} x_k \right)
\end{equation}
\begin{equation}
z_j^{(2)} = \sigma \left( \sum_{k \in N_j} \alpha_{jk} W^{(1)} z_k^{(1)} \right)
\end{equation}
where \(W^{(0)}\) and \(W^{(1)}\) are the parameters for the first and second layers of the GAT, respectively. The attention coefficients \( \alpha_{jk} \) are computed as follows:
\begin{equation}
\alpha_{jk} = \frac{\exp(\sigma(\vec{a}^T[Wx_{ij} \Vert Wx_k]))}{\sum_{r \in N_j} \exp(\sigma(\vec{a}^T[Wx_j \Vert Wx_r]))}
\end{equation}
where $z_j^{(1)}$ and $z_j^{(2)}$ are the features of vertex $j$ obtained from the first layer GAT and the second layer GAT, respectively.
For updating the adjacency matrix, it is defined as:
\begin{equation}
\hat{A} = \text{sigmoid}(Z^T \bullet Z).
\end{equation}
Next, for updating the loss function, it is defined as:
\begin{equation}
L_r^i = \text{loss}(A^i, \hat{A}^i)
\label{eq6}
\end{equation}
After feature extraction, a self-supervised clustering layer is introduced for each client. The cluster membership matrix \(U^{i} = [u_1^i, \ldots, u_k^i] \in \mathbb{R}^{K \times d_v}\) determines the soft cluster assignments \(Q^{i}\), which are calculated as follows:
\begin{equation}
q_{ij}^i = \frac{(1 + \|z_j^m - u_k^m\|^2)^{-1}}{\sum_{k} (1 + \|z_j^m - u_{k}^m\|^2)^{-1}}
\end{equation}
where $q_{ij}^i$ represents the probability that sample $j$ from the $i$-th client is assigned to cluster $k$.
Next, for client \(i\), we use the global pseudo-labels \(P\) and the soft cluster assignments \(Q^i\) to compute the KL divergence loss:
\begin{equation}
L_c^i = \text{KL}(P \Vert Q^i) = \sum_{j} \sum_{k} p_{jk} \log \frac{p_{jk}}{q_{jk}^i}
\end{equation}
Then, the total loss function is defined as:
\begin{equation}
L = L_r + \gamma L_c
\label{eq9}
\end{equation}
Here, \(L_r\) is the reconstruction loss, and \(L_c\) is the clustering loss, where \(\gamma\) is a hyper-parameter that balances the importance of the reconstruction loss relative to the clustering loss.

\subsection{Global aggregation} 
After obtaining the features \(Z^i\) of different clients and the clustering centers \(U^i\), the server uses heterogeneous aggregation on the overlapping sample features \(Z^i_C\)to obtain the global feature \(Z\), and then updates the global pseudo-labels \(P\) based on \(Z\).

To leverage the consistency and complementarity of different views, we perform heterogeneous aggregation on the overlapping sample features of different clients. Due to the heterogeneity of client data, the feature dimensions obtained are different, making direct aggregation challenging. Therefore, we assign different weights \(W_i\) to the features of different clients based on \(U^i\). Then, we concatenate the weighted overlapping sample features \(Z_C^i\) for each sample obtained from different clients to produce the high-dimensional feature \(Z\):
\begin{equation}
Z = [w_1{Z_C^1}, w_2{Z_C^{2}}, \dots, w_{m}Z_C^m]
\end{equation}
where \(w_i\) is calculated by:
\begin{equation}
w_i = 1 + \log \left( 1 + \frac{\sigma(U^i)}{\sum_v \sigma(U^i)} \right)
\label{eq11}
\end{equation}
where \( Z \in \mathbb{R}^{N \times \sum d_i} \) represents the global feature, $Z_C^{i}$ indicats the overlapping sample features of the \( i \)-th view, and $\sigma(.)$ represents the variance. Clearly, a higher variance indicates better clustering results. Therefore, by performing weighted aggregation through \( W_i \), it is possible to enhance the influence of features from views with better clustering results and reduce the influence of features from views with poorer clustering results.

The optimal cluster assignment \(C\) is then obtained using the K-means algorithm by minimizing:
\begin{equation}
\min_{C} \sum_{j=1}^{N_c} \sum_{k=1}^{K} \| z_j - c_k \|^2
\label{eq12}
\end{equation}
where \(A\) is the adjustment matrix used to align the two instances of \(S\). The server sends the global pseudo-label \(P\) to the clients as global information to integrate the complementary features of different views. The final global pseudo-labels $P$ is obtained through the following calculation:
\begin{equation}
P = \varepsilon (s_j)A,
\label{eq13}
\end{equation}
\begin{equation}
\quad s_{jk} = \frac{(1 + \| z_j - c_k \|^2)^{-1}}{\sum_j (1 + \| z_j - c_k \|^2)^{-1}}, s_{jk} \in S
\end{equation}
and
\begin{equation}
\varepsilon(s_i) = \frac{\left(\frac{s_{jk}}{\sum_j s_{jk}}\right)^2}{\sum_j \left(\frac{s_{jk}}{\sum_j s_{jk}}\right)^2}
\end{equation}
Due to differences in the cluster centers in each round of aggregation, the Hungarian algorithm is used to introduce $A$ to align $S$ across different communication rounds.
Finally, the most likely cluster assignment for each data point is determined by:
\begin{equation}
y_j = \arg \max_{jk} \left(\frac{1}{m}{\sum_m q_{jk}^i}\right)
\end{equation}

\subsection{Algorithm optimization}
 As shown in Algorithm 1, the optimization in the proposed FIM-GNNs primarily consists of the client and server components. 
 Firstly, we carry out the pre-training of the heterogeneous GNNs at the client side. We use all the data \(X^i\) from each view, and train the encoder and decoder only using the reconstruction loss to enhance the model's capability in feature extraction. In the proposed FIM-GNNs, we initialize \(k\) cluster centers using the K-means algorithm. Subsequently, we train the FIM-GNNs using the holistic federated framework, where the server aggregates the heterogeneous features \(Z_C^i\) of overlapping samples from each client into a global feature \(Z\), based on which global pseudo-labels \(P\) are generated and sent back to the clients. Each client then uses the overlapping sample data \(X_C^i\) and the global pseudo-labels \(P\) to train the model in parallel for \(T\) rounds, obtaining features \(Z_C^i\) and cluster centers \(U_i\). The server and clients update alternately over \(E\) communication rounds.
\begin{algorithm}
\caption{Optimization process in the proposed FIM-GNNs}
\begin{algorithmic}[1]
\State \textbf{Input:} Datasets $X = \{X^1, \ldots, X^m\}$, adjacency matrices $A = \{A^1, \ldots, A^m\}$, number of clusters $K$, number of communication rounds $E$, number of training rounds $T$.
\State \textbf{Output:} Cluster results
\While{not reaching $E$}
    \For{$m = 1$ to $M$ in parallel}
        \If{$E = 1$}
            \State Pretrain the autoencoders by Eq. (\ref{eq6}).
        \Else
            \While{not reach $T$}
                \State Update $Z_m, U_m$ by optimizing Eq. (\ref{eq9})
            \EndWhile
        \EndIf
        \State Upload $Z_m$ and $U_m$ to the server.
    \EndFor
    \State Update $Z$ by Eq.  (\ref{eq11}).
    \State Update $C$ by Eq.  (\ref{eq12}).
    \State Obtain $P$ by Eq.  (\ref{eq13}).
    \State Distribute $P$ to each client.
\EndWhile
\State Calculate the clustering predictions based on Eq. (13).
\end{algorithmic}
\end{algorithm}

\subsection{Complexity Analysis}
Algorithm 1 consists of two main phases, which together determines its computational complexity. The algorithm begins with training each view separately. This step has a complexity of \(O(N(dd_1 + d_1d_2))\), where \(N\) and \(d\) represent the sample size and sample dimension, respectively.  \(d_1\) and \(d_2\) are the feature dimensions of the views. Next, in the aggregation phase, each client contributes to the global model by calculating the feature matrix \(P\) and cluster centers \(C\), resulting in a complexity of \(O(NDK)\), where \(D\) is the feature dimension, and \(K\) is the number of clusters. For updating \(P\), the complexity is \(O(K^3 + NK)\). Therefore, the proposed FIM-GNNs is with a complexity of \(O(K^3 + NDK + N(dd_1 + d_1d_2))\).

\section{Experiments}
\subsection{Experimental Setup}

Based on the differences in sample features across various views within the federated framework, we conducted training on two widely used datasets, Caltech-7 and BDGP, to validate the performance of our proposed FIM-GNNs. Table \ref{dd} show the description of the datasets. In the experiments, we construct the graph structure information as in \cite{xiao2023dual}.
Taking into account the differences in dimensionality and quantity of data across different clients, we select partial views as the dataset. Simultaneously, based on the specific missing rates $R_{i}$ of each view, parts of the samples are missing, but we ensure that each sample exists in at least one view.  %Firstly, we construct a KNN graph based on the Euclidean distance between two points. Then, we adopt two pruning strategies to improve the quality of the graph.

\begin{table}[ht] \small
\centering
\begin{tabular}{@{} l|cccl @{}}
\toprule
Datasets   & $C$ & $N$                      & $D_i$ \\ \hline
%BBCSport   & 5 & 544  & 489,272              & 3,183,3203 \\
Caltech-7  & 7 & 1474     & 928,512,254 \\
%Out-Scene  & 8 & 2688 & 2,150,2,150,2,419    & 512,432,256 \\
BDGP       & 5 & 2000     & 1,000,500,250 \\
\bottomrule
\end{tabular}
\caption{Dataset Description}
\label{dd}
\end{table}

We firstly use two standard clustering metrics, namely accuracy (ACC) and Normalized Mutual Information (NMI) for evaluations. Additionally, we use the Adjusted Rand Index (ARI) to assess the quality of clustering. We compared the proposed FIM-GNNs with four state-of-the-art incomplete multi-view clustering methods. CDIMC-NET \cite{Jie2021} is a method that integrates deep encoders and graph embedding strategies, and introduces a self-paced strategy to select optimal samples for model training. COMPLETER \cite{lin2021completer} is a contrastive learning-based method that achieves cross-view data recovery from an information-theoretic perspective. APADC \cite{xu2023adaptive} is an adaptive feature projection-based incomplete multi-view clustering method, which introduces distribution-aligned feature learning. IMVC-CBG \cite{wang2022highly} is an anchor-based multi-view learning method, introducing a bipartite graph framework to address incomplete multi-view clustering.

In our experiments, we set the number of epochs to 10, with a learning rate initialized at 0.005 and reduced by half every 50 epochs. We use Adam optimizer. % For the models BBCSport, Out-Scene, and BDGP, we apply the GCN network, setting initial learning rates specifically tailored for each dataset. For Caltech-7, we utilize a different configuration, adjusting learning rates and epochs accordingly.
For both the GAT and GCN frameworks, the Adam optimizer is used, with a pre-training learning rate of 0.005 and a training learning rate of 0.001. For the GAT framework, on the BDGP dataset, the dimensions of the two GAT layers are [128,16], and on the Caltech-7 dataset, the dimensions of the two layers are [512,32]. For the GCN framework, the dimensions of the two GCN layers are [128,16]. When the missing rate $\beta$ does not exceed 0.1, the GCN framework is used; when the missing rate exceeds 0.1, the GAT framework is used. The hyperparameter $\gamma$ is set to 1.
\begin{table*}[h]
\centering
\setlength{\tabcolsep}{9pt}
\scalebox{0.85}{
\begin{tabular}{cl|ccc|ccc|ccc} \toprule
 &  [$R_1$,$R_2$,$R_3$]& \multicolumn{3}{c|}{{[0.2,0.05,0.05]}} & \multicolumn{3}{c|}{{[0.2,0.2,0.1]}} & \multicolumn{3}{c}{{[0.3,0.3,0.1]}} \\ \cline{2-11} 
 {Datasets} & {Metrics} & {ACC} & {NMI} & {ARI} & {ACC} & {NMI} & {ARI} & {ACC} & {NMI} & {ARI}\\ \hline
\multirow{5}{*}{Caltech-7} & CDIMC-NET & 0.532 & \textbf{0.564} & \textbf{0.419} & 0.362 & 0.114 & 0.084 & 0.509 & \textbf{0.512} & 0.371 \\
 & COMPLETER & 0.476 & 0.342 & 0.348 & 0.579 & 0.469 & 0.332 & 0.476 & 0.345 & 0.342 \\
 & APADC & 0.544 & 0.542 & 0.403 & 0.462 & 0.413 & 0.232 & 0.534 & 0.411 & 0.384 \\
 & IMVC-CBG & 0.502 & 0.351 & 0.284 & 0.513 & 0.298 & 0.224 & 0.521 & 0.291 & 0.203 \\
 & FIM-GNNs & \textbf{0.604} & 0.426 & 0.348 & \textbf{0.582} & \textbf{0.489} & \textbf{0.397} & \textbf{0.565} & 0.479 & \textbf{0.391} \\ \hline
\multirow{5}{*}{BDGP} & CDIMC-NET & 0.369 & 0.107 & 0.008 & 0.368 & 0.161 & 0.113 & 0.367 & 0.101 & 0.009 \\
 & COMPLETER & 0.341 & 0.105 & 0.103 & 0.357 & 0.113 & 0.084 & 0.345 & 0.106 & 0.104 \\
 & APADC & 0.291 & 0.135 & 0.041 & 0.283 & 0.032 & 0.031 & 0.266 & 0.091 & 0.021 \\
 & IMVC-CBG & 0.521 & 0.303 & 0.268 & 0.501 & 0.293 & 0.258 & 0.491 & 0.287 & 0.261 \\
 & FIM-GNNs & \textbf{0.614} & \textbf{0.426} & \textbf{0.399} & \textbf{0.625} & \textbf{0.470} & \textbf{0.435} & \textbf{0.580} & \textbf{0.368} & \textbf{0.331}  \\ \bottomrule
\end{tabular}}
\caption{Performance comparison of different missing rates on Caltech-7 and BDGP Datasets}
\label{tab3}
\end{table*}
\subsection{Performance Evaluation}
Table \ref{tab3} summarizes the experimental results on two datasets, where $[R_{1},R_{2},R_{3}]$ means the missing rates on different clients.It can be observed that our method basically achieves the best results in five incomplete multi-view clustering methods, demonstrating the effectiveness of our approach. Compared with three deep neural network-based methods (CDIMC-NET, COMPLETER and APADC), the FIM-GNNs achieves better results, indicating that the introduction of heterogeneous graph structure as auxiliary information can effectively improve model performance. Particularly on the BBCSport dataset, our method outperforms the other three deep methods significantly, suggesting that the simultaneous use of heterogeneous graph structure and global node feature information can more effectively capture features, especially in federated settings, thereby improving clustering results for incomplete view data in different scenarios.

\begin{figure}[htp]
    \centering
    \includegraphics[width=8cm]{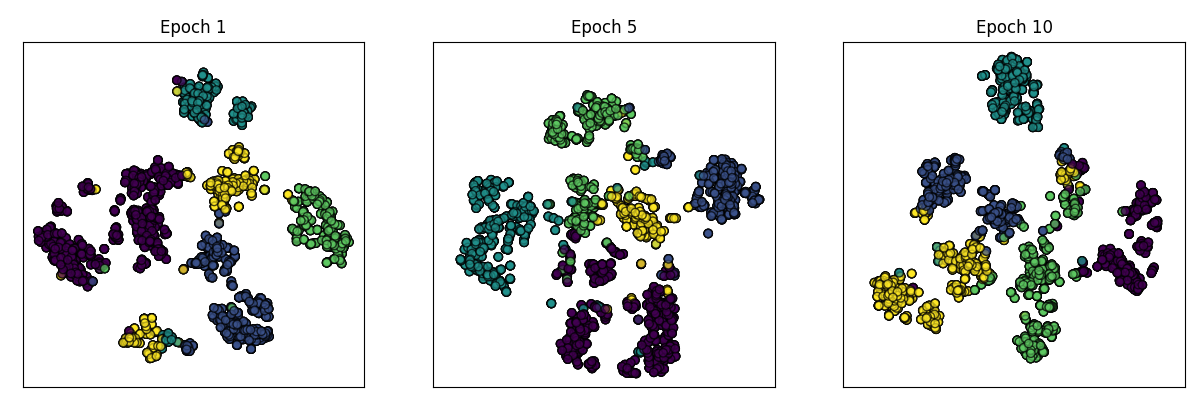}
    \caption{The $t$-SNE visualization results for different communication epochs on the BDGP dataset.}
    \label{arch13}
\end{figure}

In addition, to visually observe the clustering results more intuitively, we utilized $t$-SNE for visualization of the results with a missing rate of [0.2,0.2,0.1] on the BDGP dataset. 
Figure \ref{arch13} presents the visualization results after reducing the features of complete samples to two-dimensional, where different colors represent different clusters. It can be observed that initially, the nodes are relatively scattered, and the boundaries between different clusters are not distinct. As the communication rounds increase, nodes within the same cluster gradually aggregate, and the boundaries between different clusters become clearer.
This is because a global pseudo-label mechanism with heterogeneous aggregation is beneficial for obtaining the accuracy of the federated incomplete multi-view clustering results. 

\subsection{Effect of heterogeneous GNNs}
\begin{figure}[htp]
    \centering
    \includegraphics[width=8cm]{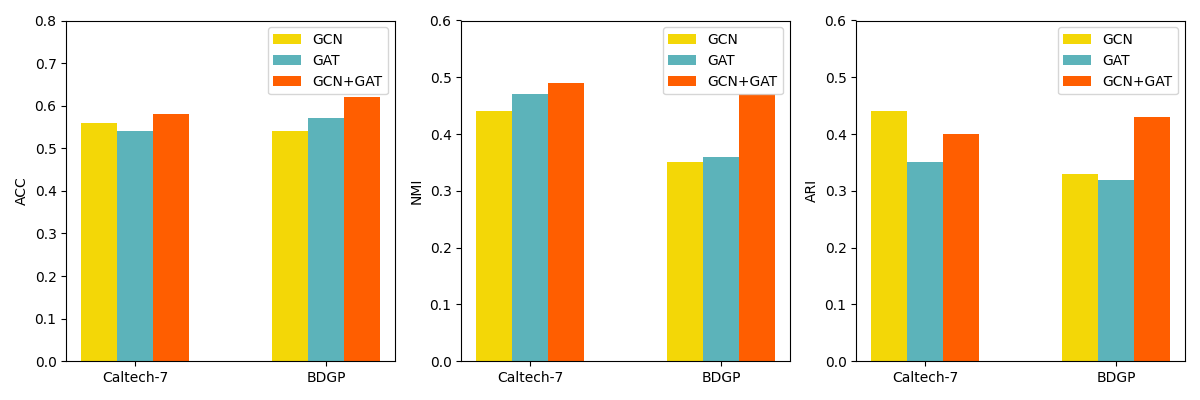}
    \caption{Experimental results of the FIM-GNNs with GAT, GCN, and a combination of GCN and GAT, respectively.}
    \label{arch111}
\end{figure}

To evaluate the impact of heterogeneous GNNs, we assessed the performance of FIM-GNNs with GAT, GCN, and a combination of GCN and GAT, respectively. As shown in Fig. \ref{arch111}, the experiments across two datasets indicated that the combination of GCN and GAT generally yields better results, while the individual models (GAT and GCN) also perform well. The performance of the combination of GCN and GAT demonstrates the effectiveness of heterogeneous GNNs in dealing with the incompleteness of performance across different views.

\subsection{Parameter Sensitivity}
\begin{figure}[ht]
    \centering
    \includegraphics[width=9cm]{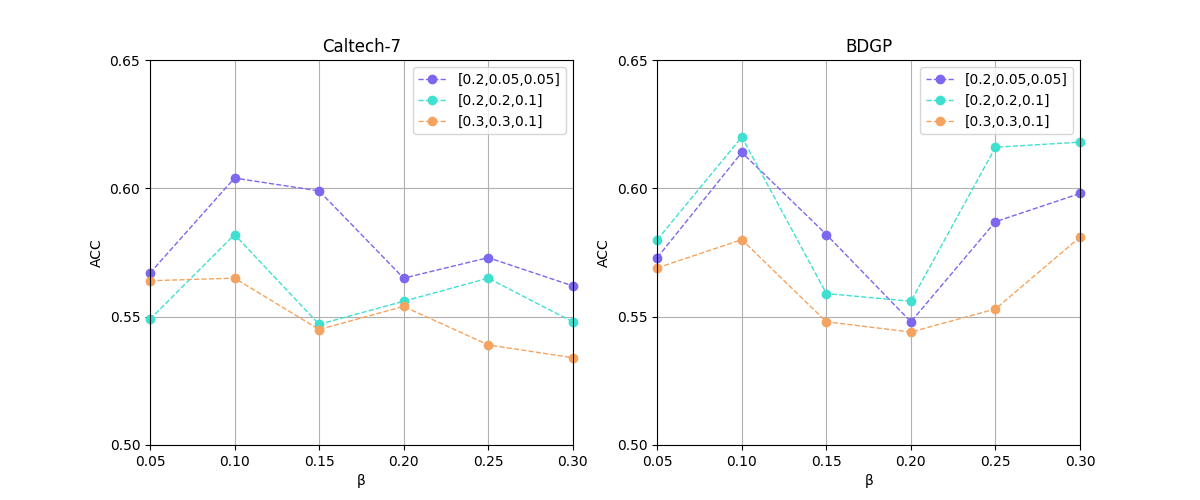}
    \caption{Accuracy results were obtained by using different $\beta$ values on the Caltech-7 and BDGP datasets.}
    \label{arch16}
\end{figure}

We determine the model for each client based on the missing rate threshold $\beta$ in the each view. When the view missing rate does not exceed $\beta$, we use GCN; otherwise we use GAT. To verify the effectiveness of $\beta$, we run the FIM-GNNs with $\beta$ set at [0.05, 0.10, 0.15, 0.20, 0.25, 0.30], and the results are shown in the Fig. \ref{arch16}. We can see from the observations that $ \beta = 0.1$ yields the best performance on Caltech-7 and BDGP datasets.

\begin{figure}[htp]
    \centering
    \includegraphics[width=8cm]{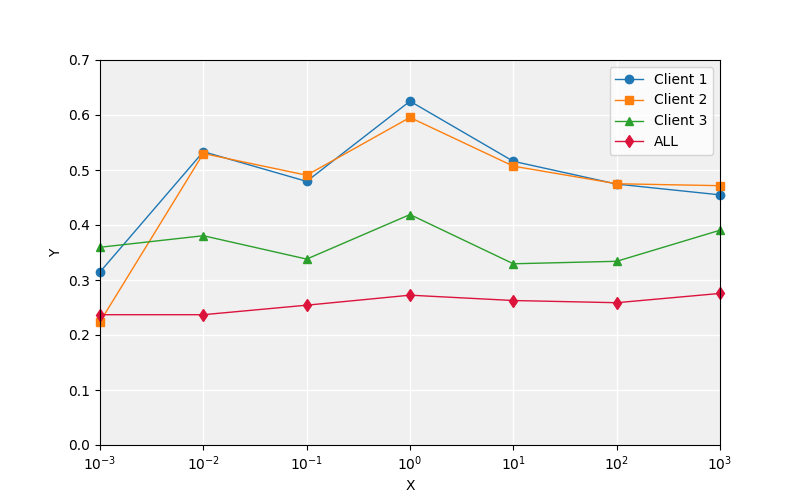}
    \caption{Accuracy results obtained using different $\gamma$ values on the BDGP dataset, ranging from $10^{-3}$ to $10^{3}$.}
    \label{arch15}
\end{figure}

During local training on the client side, a hyperparameter $\gamma$ is used to balance the clustering loss and reconstruction loss. We conducted experiments on the BDGP dataset by varying the values of the hyperparameter $\gamma$ from $10^{-3}$, $10^{-2}$, ..., $10^{2}$, $10^{3}$ to test parameter sensitivity, as shown in Fig. \ref{arch15}. It can be observed that different clients exhibit varying sensitivities to $\gamma$, but overall, the performance is optimal when $\gamma=1$, so we set $\gamma=1$ in this study.

\section{Conclusion}
In this study, we present a federated incomplete multi-view clustering framework with heterogeneous GNNs. Local client-side training and global heterogeneous aggregation are introduced to effectively enhance the performance of federated incomplete multi-view clustering. Moreover, we employ a global pseudo-label mechanism with heterogeneous aggregation to deal with incomplete view data. The effectiveness of FIM-GNNs is evaluated on public datasets compared to state-of-the-art incomplete multi-view clustering methods. Although BGNAS can obtain effective and efficient clustering results for incomplete multi-view data, it is still worth considering its applicability to extremely imbalanced multi-view data or other complex clustering tasks. In the future, the proposed FIM-GNNs can potentially be integrated with heterogeneous GNNs for incomplete multi-modal clustering tasks or specific domains.

\appendix

\section*{Ethical Statement}

There are no ethical issues.

\section*{Acknowledgments}
 This work was supported in part by the National Natural Science Foundation of China under Grant No. 62136003, and in part by the Guangdong Basic and Applied Basic Research Foundation under Grant No. 2024A1515011729 and No. 2023A1515012534.

\bibliographystyle{named}
\bibliography{ijcai23}

\end{document}